\documentclass[11pt,a4paper]{article}
\usepackage[hyperref]{acl2018}
\usepackage{times}
\usepackage{latexsym}
\usepackage{amssymb}

\usepackage{graphicx}
\usepackage[linesnumbered,ruled]{algorithm2e}

\usepackage{mathtools}

\usepackage{url}

\aclfinalcopy 

\title{Correcting the Common Discourse Bias in Linear Representation of Sentences using Conceptors}

\author{Tianlin Liu\thanks{Research done while visiting University of Pennsylvania.} \\
  Department of Computer Science and \\ 
  Electrical Engineering\\
Jacobs University Bremen \\
28759 Bremen, Germany\\
\texttt{t.liu@jacobs-university.de} \\
  \\\And
  Jo\~{a}o Sedoc \and Lyle Ungar \\
Department of Computer and \\  Information Science\\
University of Pennsylvania\\
Philadelphia, PA 19104\\
\texttt{\{joao,ungar\}@cis.upenn.edu}}

\date{}

\begin{document}
\maketitle

\begin{abstract}
Distributed representations of words, better known as {\em word embeddings}, have become important building blocks for natural language processing tasks. Numerous studies are devoted to transferring the success of unsupervised word embeddings to {\em sentence embeddings}. In this paper, we introduce a simple representation of sentences in which a sentence embedding is represented as a weighted average of word vectors followed by a soft projection. We demonstrate the effectiveness of this proposed method on the clinical semantic textual similarity task of the BioCreative/OHNLP Challenge 2018. 
\end{abstract}

\section{Introduction}

The success of unsupervised word embeddings has motivated researchers to learn embeddings for larger chunks of text such as sentences. Current research in sentence embedding is mainly advancing along two lines. In one line, researchers use powerful and complex models such as deep neural networks and recurrent neural networks to capture the semantics of sentences \citep{Blunsom2014, Iyyer2015, Yin2015, Cer2018}. In a complementary second line, researchers have invented computationally cheap alternatives to embed sentences using simple linear algebraic operations \citep{Wieting2016, Arora2017, Mu2017b, Khodak2018, Ethayarajh2018}. Surprisingly, many simple methods yield comparable or even better results compared to complicated methods, particularly in out-of-domain tasks \citep{Wieting2016}. The current paper follows the second avenue of research.

Among all methods for sentence embedding, arguably the simplest one is to compute a sentence embedding as the average of the sentence's word vectors. This naive approach has proven to be a formidable baseline for many downstream natural language processing (NLP) tasks \citep{Faruqui2014, Wieting2016}.  However, it comes with a limitation: Since word vectors of a given sentence are spanned by a few leading directions \citep{Mu2017,Khodak2018}, averaging the word vectors amplify these leading directions while diminishing the useful signals contained in the trailing directions. We refer this problem as the \emph{common direction bias} in linear representations of sentences.

To correct the common direction bias, researchers have invented a ``common component removal'' trick \citep{Arora2017, Mu2018}. This technique removes the top one or top few principal components from the word vectors \citep{Mu2018} or the weighted average thereof \citep{Arora2017}. Intuitively, since dominating directions of word vectors tend to influence the additive composition in the same way, nulling out such directions ameliorates the effect. Post-processed with such a technique, linear representations of sentences usually deliver strong performances, sometimes outperforming sophisticated Deep-Learning based methods including RNN's and LSTM's on standard benchmarks \citep{Arora2017, Mu2018, Ethayarajh2018}.

Although common component removal has proven to be effective, the technique is liable either to not remove enough noise or to cause too much information loss \citep{Khodak2018}. In this paper, we propose a novel and simple way to address this issue. Our proposed method can be regarded as a ``soft'' version of common component removal. Specifically, given a sequence of word vectors, we softly down-weight principal components (PCs) with the assistance of a regularized identity map called a \emph{Conceptor} \citep{Jaeger2017}.

The rest of the paper is organized as follows. We first review the linear representation of sentences by \citet{Arora2017}. We then introduce the Conceptor approach for soft common component removal, which is the main contribution of this paper. After that, we demonstrate the effectiveness of the proposed method on the Clinical STS dataset of the BioCreative/OHNLP Challenge 2018.  

\section{Linear representation of sentences}
We present a brief sketch of the sentence embedding method by \citep{Arora2017}. In \citep{Arora2017}, words in a sentence are assumed to be generated by a ``discourse'' $c_s$,  which is a vector-valued random variable taking values in $\mathbb{R}^N$. \citet{Arora2017} further assume that there exists a fixed common discourse vector $c_0$ which is orthogonal to all realizations of $c_s$, i.e., $c_0 \bot c_s$. Given a discourse $c_s$, the emitting probability for a word $w$ is assumed to be

\begin{align*}
p ( w \,|\, c_s) = \gamma p(w) + (1- \gamma )\frac{\exp( \tilde{c}_s^\top v_w)}{Z_{\tilde{c}_s}} ,
\end{align*}
where $v_w$ is the word vector for the word $w$, $p(w)$ is the monogram probability for the word $w$, $Z_{\tilde{c}_s} := \sum_{w \in V} \exp(\tilde{c}_s^\top v_w)$ is the normalizing term, $\tilde{c}_s := \beta c_0 + (1-\beta)c_s$, and $\gamma$, $\beta$ are scalar hyper-parameters. As a result, this model favors to produce two types of words: those words with high monogram probability and those words whose vector representation located close to both $c_0$ and $c_s$ (up to a balancing parameter $\beta$). Using this model, \cite{Arora2017} derived a sentence algorithm which contains two steps. In the first step, $\tilde{c}_s$ is approximated by $\hat{\tilde{c}}_s$, which has the form 

\begin{equation} \label{eq:sif1}
 \hat{\tilde{c}}_s \coloneqq \frac{1}{ |s|}\sum_{w \in s} \frac{a}{p(w) + a} v_w 
 \end{equation}
for a scalar hyper-parameter $a$; in the second step, the common discourse $c_0$ is estimated as the first PC $u$ of a set of sentences $\{\hat{\tilde{c}}_s\}$ via an uncentered PCA. The final sentence embedding $\hat{c}_s$ is consequently obtained by removing its projection on the first PC, i.e., by letting 

\begin{equation} \label{eq:sif2}
\hat{c}_s  \coloneqq  \hat{\tilde{c}}_s - u u^{\top} \hat{\tilde{c}}_s 
\end{equation}
as an approximation of $c_s$.

We now take a more abstract view on the common component removing step, i.e., Equation \ref{eq:sif2}. This step relies on a key assumption: There exists a \emph{single direction} $c_0$ which represents a syntax (i.e. function word)-related ``discourse''. As a straightforward generalization, one can also assume that there exists a proper $d$-dimensional \emph{linear subspace} $\mathcal{M} \subset \mathbb{R}^d$, where $ d < N$, such that all discourses $c'_0 \in \mathcal{M}$ are syntax-related discourses. Under this assumption, one can define a projection matrix $\mathbf{P}_{\mathcal{M}}: \mathbb{R}^N \to \mathcal{M}$ which characterizes the subspace of common discourses. To separate $\hat{c}_s$ from $\hat{\tilde{c}}_s$, one projects $\hat{\tilde{c}}_s$ to the orthogonal complement of $\mathcal{M}$, written as $\mathcal{M}^{\perp}$, by letting $\hat{c}_s \coloneqq \mathbf{P}_{\mathcal{M}^{\perp}} \hat{\tilde{c}}_s = (\mathbf{I} - \mathbf{P}_{\mathcal{M}} ) \hat{\tilde{c}}_s = \hat{\tilde{c}}_s - \mathbf{P}_{\mathcal{M}} \hat{\tilde{c}}_s$, where $\mathbf{I}$ is an identity matrix. In particular, choosing $\mathbf{P}_{\mathcal{M}}  = u u^\top$, where $u$ is the first PC of a set of sentences $\{\tilde{c}_s\}$, we recover the second step of \citep{Arora2017}. As an alternative, we can also choose $\mathbf{P}_{\mathcal{M}}  = U_{1:D} U_{1:D}^\top$, where $U_{1:D}$ is a matrix whose columns are the first $D$ PCs of a set of sentences or a set of words. This alternative has been investigated by \citet{Mu2018}. 

As shown in \citet{Mu2018}, the number $D$ plays a crucial role in the effect of common component(s) removal. In many situations, a fixed integer $D$ makes this approach liable to either not remove enough noise or to cause too much information loss \citep{Khodak2018}. We therefore propose an alternative method which removes the common components in a ``softer'' manner.

Our starting point is a relaxation of the key assumptions of \citet{Arora2017} and \citet{Mu2018}. Instead of assuming that the function words to be allocated along a single direction \citep{Arora2017} or to be constrained in a proper linear subspace \citep{Mu2018}, we allow function and common words to span the whole $\mathbb{R}^{N}$. This assumption admits a more realistic modeling. Indeed, we find that the word vectors of stop words \citep{Stone2011} span the entire space of $\mathbb{R}^{N}$.  

Allowing function words to span the whole $\mathbb{R}^{N}$, however, leads an obstacle: We can not project the sentence embedding to the orthogonal complement of such a space.  To address this issue, we use the Conceptor matrix \citep{Jaeger2017} to approximate the space occupied by function words. 

\section{Conceptors as soft subspace projection maps}

In this section we briefly introduce matrix Conceptors, sometimes using the wordings of \citet{Jaeger2017}. Consider a set of vectors  $\{x_1, \cdots, x_n\}$,  $x_i \in \mathbb{R}^N$ for all $i \in \{1, \cdots, n\}$. A Conceptor matrix (under the assumption that data points $\{x_1, \cdots, x_n\}$ are identically distributed) can be defined as a regularized identity map $C$ that minimizes
\begin{equation}
\label{matconceptor}
\frac{1}{n} \sum_{i=1}^{n}  \|x_i - C x_i\|_2^2+\alpha^{-2}\|C\|_{\text{F}}^2.
\end{equation}
where $\| \cdot \|_{\text{F}}$ is the Frobenius norm and $\alpha^{-2}$ is a scalar parameter called \emph{aperture}.

It can be shown that $C$ has a closed form solution: 
\begin{equation}
\label{conceptorsolution}
C = \frac{1}{n} X X^{\top}(\frac{1}{n} XX^{\top}+\alpha^{-2} I)^{-1},
\end{equation}
where $X$ is a $N\times n$ data collection matrix whose $i$-th column is $x_i$. Assuming that the singular value decomposition (SVD) of $\frac{1}{n} X X^{\top}$ has the form $\frac{1}{n} X X^{\top} = U \Sigma U^{\top}$, we can re-write $C$ as s $USU^\top$, where singular values $s_i$ of $C$ can be written in terms of the singular values $\sigma_i$ of $R$: $s_i = \sigma_i / (\sigma_i + \alpha^{-2}) \in [0, 1)$.  Applying Conceptors on the averaged word vectors $\{\hat{\tilde{c}}_s\}$, i.e., using $\{\hat{\tilde{c}}_s\}$ in place of $\{x_1, \cdots, x_n\}$ in Equation \ref{matconceptor}, we see that the columns of the matrix $U$ are exactly the PCs estimated via the un-centered PCA of $\{\hat{\tilde{c}}_s\}$. In particular, the first column of $U$ is the first PC $u$ used in Equation \ref{eq:sif2} by \citet{Arora2017} introduced in the previous section.

We now study a matrix $G \coloneqq \textbf{I} - C$. This matrix characterizes a linear subspace that can be roughly understood as the orthogonal complement of the subspace characterized by $C$. This fact can be seen via the following representation: 

\begin{align}
\label{SVDC2}
G \coloneqq \textbf{I} - C & = U
\begin{bmatrix}
\frac{\alpha^{-2}}{\sigma_{1}+\alpha^{-2}} & & \\
& \ddots & \\
& & \frac{\alpha^{-2}}{\sigma_{N}+\alpha^{-2}}
\end{bmatrix} U^\top. 
\end{align}

Note that $G$ can be considered as a soft projection matrix which down-weights the leading PCs of $\{\hat{\tilde{c}}_s\}$: For vectors $\hat{\tilde{c}}_s$ in the linear subspace spanned by the leading PCs with large variance, $G \hat{\tilde{c}}_s \approx 0$; for vectors $\hat{\tilde{c}}_s$ in the linear subspace spanned by trailing PCs with low variance, $G \hat{\tilde{c}}_s \approx \hat{\tilde{c}}_s$. The \emph{soft} projection $G \hat{\tilde{c}}_s$ has the following relationship with the \emph{hard} projection in Equation \ref{eq:sif2}: It is clear that, if we modify $G$ into $G'$, where $G' = U \text{diag}([0, 1, 1, \cdots, 1 ]) U^\top$ (c.f. Equation \ref{SVDC2}), we recover the result in Equation \ref{eq:sif2}:
\begin{eqnarray*}
G' \hat{\tilde{c}}_s & = & U \text{diag}([0, 1, 1, \cdots, 1 ]) U^\top \hat{\tilde{c}}_s  \\
& = & \hat{\tilde{c}}_s - u u^{\top} \hat{\tilde{c}}_s.
\end{eqnarray*}

Besides applying Conceptors on the averaged word vectors $\{\hat{\tilde{c}}_s\}$, another reasonable approach is to directly apply Conceptors $C$ on all word vectors $\{v_w \}$ which constituent sentences in a dataset. The Conceptors learned in this way have a more transparent interpretation: they characterize the shared linear subspace of mainly two types of words (i) function words that have little lexical meaning, (ii) frequent but non-function words in a particular dataset, which can be regarded as a shared background information of a dataset. In practice, we find that learning Conceptors directly from words vectors usually delivers better results than from averaged word vectors, and therefore we use the former method throughout the numerical experiments presented below. 

To help $C$ capture the space spanned by the two types of words introduced previously, we find that we obtain better results if we estimate $C$ not only based on word vectors appearing in the set of actual sentences but also based on the word vectors appearing in a predefined set of stop words. Such a set of stop words can be thought of a prior which describes the subspace of common words. The overall sentence embedding procedure is displayed in Algorithm \ref{algo:conceptorAlg}.

\begin{algorithm}[!h]
\SetKwInOut{Input}{Input}
\SetKwInOut{Output}{Output}
\Input{Word embeddings $\{v_w: w \in V\}$, a set of sentences $\mathcal{S}$, parameter $a$,  parameter $\alpha$, a set of estimated probabilities $\{p(w) : w \in V\}$ of the words, and a set of stop words \text{Q}.}
\For{$s \in \mathcal{S}$}{%
$v_s \leftarrow \frac{1}{s} \sum_{w \in s} \frac{a}{ a + p(w)} v_w$
      }
Form a matrix $X$ whose columns are word vectors from the set $\{v_w, \forall w \in s, s \in S \} \cup \{v_{w'} , \forall v_{w'} \in Q \}$ and calculate the Conceptors $G$ based on Equation \ref{matconceptor} and \ref{SVDC2}.\\
\For{$s \in \mathcal{S}$}{%
$v_s \leftarrow G v_s $
}
\Output{Sentence embeddings $\{v_s\}$.}
\caption{Sentence Embedding with Conceptors.}
\label{algo:conceptorAlg}
\end{algorithm}
\section{Evaluation}

       We apply our proposed method to the BioCreative/OHNLP Challenge 2018 \citep{Wang2018}. Similar to the SemEval STS challenge series \citep{Cer2017}, the BioCreative/OHNLP Challenge 2018 offers a platform to evaluate the semantic similarity between a pair of sentences and compare the results with manual annotations. Constructing a dataset by gathering naturally occurring pairs of sentences in the clinical context is a challenging task on its own. For the detailed description of the dataset, we refer the readers to \citep{Wang2018}.

\begin{figure}[h]
\includegraphics[scale = 0.38]{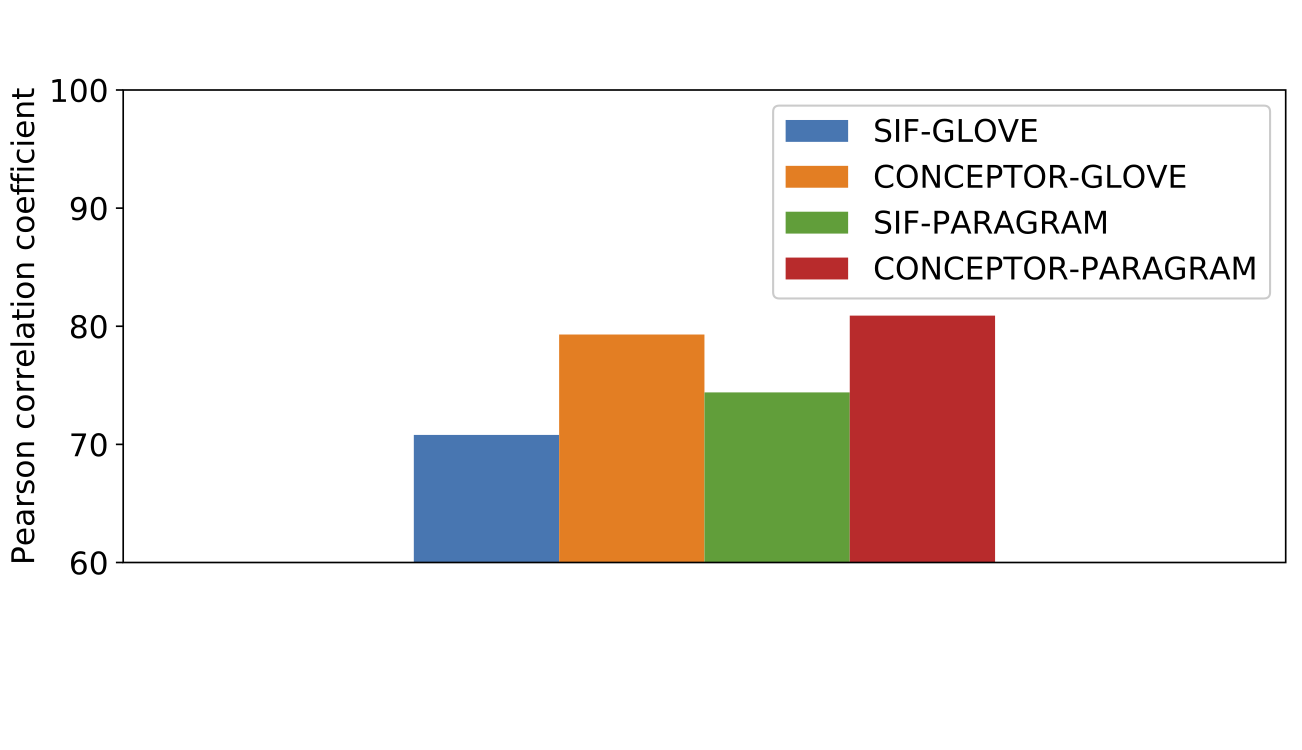}
\caption{Experimental results on the training dataset. \label{fig:sts_result}}
\end{figure}

For preprocessing, we use the \texttt{nltk} Python package \citep{Bird2009} to tokenize the words in the sentences. We discard all punctuations. To estimate the monogram probabilities of words, we use word frequencies collected from Wikipedia\footnote{\url{https://github.com/PrincetonML/SIF/blob/master/auxiliary\_data/enwiki\_vocab\_min200.txt}}. We use two sets of pretrained word vectors, GloVe \citep{Pennington2014} and Paragram-SL999 \citep{Wieting2015}.  For the hyper-parameters in Algorithm \ref{algo:conceptorAlg}, we use the set of stop words collected by \citet{Stone2011}; we fix the aperture $\alpha^{-2} = 1$ for the experiments; we choose $a = 0.001$ as done in \citet{Arora2017}. The experimental results in the metric of Spearman's rank correlation coefficient for sentence similarities are shown the figure \ref{fig:sts_result}, where the similarity between two sentence vectors is evaluated using cosine distance. 

We see that the sentence embeddings processed with Conceptors (red and orange bars in Figure \ref{fig:sts_result}) favorably outperform the sentences processed with \emph{Smooth Inverse Frequency} (SIF) \citep{Arora2017} approach, which removes the first PC of the set of sentences in the training dataset.

\section{Conclusion}

In this paper, we described how to use a regularized identity map named Conceptors to correct the common component bias in linear sentence embedding. The goal is to softly project the sentence embeddings away from the principal components of word vectors which correspond to high variances. Empirically, we find the proposed method outperforms the baseline method of \citep{Arora2017}. In future work, we will combine this method with the recently proposed unsupervised random walk sentence embedding \citep{Ethayarajh2018}.

\section*{Acknoledgements}
The authors thank anonymous reviewers for their helpful comments. Tianlin Liu appreciates a travel grant of BioCreative/OHNLP Challenge 2018 at the ACM-BCB 2018 (grant number 5R01GM080646-12).

\bibliographystyle{acl_natbib}
\bibliography{/Users/liutianlin/Desktop/Academics/NLP/nlp_progress.bib}

\begin{thebibliography}{19}
\expandafter\ifx\csname natexlab\endcsname\relax\def\natexlab#1{#1}\fi

\bibitem[{Arora et~al.(2017)Arora, Liang, and Ma}]{Arora2017}
S.~Arora, Y.~Liang, and T.~Ma. 2017.
\newblock A simple but tough-to-beat baseline for sentence embeddings.
\newblock In \emph{International Conference on Learning Representations}.

\bibitem[{Bird et~al.(2009)Bird, Klein, and Loper}]{Bird2009}
S.~Bird, E.~Klein, and E.~Loper. 2009.
\newblock \emph{Natural Language Processing with Python}, 1st edition.
\newblock O'Reilly Media, Inc.

\bibitem[{Blunsom et~al.(2014)Blunsom, Grefenstette, and
  Kalchbrenner}]{Blunsom2014}
P.~Blunsom, E.~Grefenstette, and N.~Kalchbrenner. 2014.
\newblock A convolutional neural network for modelling sentences.
\newblock In \emph{Proceedings of the 52nd Annual Meeting of the Association
  for Computational Linguistics}. Proceedings of the 52nd Annual Meeting of the
  Association for Computational Linguistics.

\bibitem[{Cer et~al.(2017)Cer, Diab, Agirre, Lopez-Gazpio, and
  Specia}]{Cer2017}
D.~Cer, M.~Diab, E.~Agirre, I.~Lopez-Gazpio, and L.~Specia. 2017.
\newblock Semeval-2017 task 1: Semantic textual similarity-multilingual and
  cross-lingual focused evaluation.
\newblock \emph{arXiv preprint arXiv:1708.00055}.

\bibitem[{Cer et~al.(2018)Cer, Yang, Kong, Hua, Limtiaco, John, Constant,
  Guajardo-Cespedes, Yuan, Tar et~al.}]{Cer2018}
D.~Cer, Y.~Yang, S.~Kong, N.~Hua, N.~Limtiaco, R.~John, N.~Constant,
  M.~Guajardo-Cespedes, S.~Yuan, C.~Tar, et~al. 2018.
\newblock Universal sentence encoder.
\newblock \emph{arXiv preprint arXiv:1803.11175}.

\bibitem[{Ethayarajh(2018)}]{Ethayarajh2018}
K.~Ethayarajh. 2018.
\newblock Unsupervised random walk sentence embeddings: A strong but simple
  baseline.
\newblock In \emph{Proceedings of The Third Workshop on Representation Learning
  for NLP}, pages 91--100. Association for Computational Linguistics.

\bibitem[{Faruqui et~al.(2014)Faruqui, Dodge, Jauhar, Dyer, Hovy, and
  Smith}]{Faruqui2014}
M.~Faruqui, J.~Dodge, S.~K. Jauhar, C.~Dyer, E.~Hovy, and N.~A. Smith. 2014.
\newblock Retrofitting word vectors to semantic lexicons.
\newblock \emph{arXiv preprint arXiv:1411.4166}.

\bibitem[{Iyyer et~al.(2015)Iyyer, Manjunatha, Boyd-Graber, and H}]{Iyyer2015}
M.~Iyyer, V.~Manjunatha, J.~Boyd-Graber, and Daum{\'e}~III H. 2015.
\newblock Deep unordered composition rivals syntactic methods for text
  classification.
\newblock In \emph{Proceedings of the 53rd Annual Meeting of the Association
  for Computational Linguistics and the 7th International Joint Conference on
  Natural Language Processing (Volume 1: Long Papers)}, volume~1, pages
  1681--1691.

\bibitem[{Jaeger(2017)}]{Jaeger2017}
H.~Jaeger. 2017.
\newblock Using conceptors to manage neural long-term memories for temporal
  patterns.
\newblock \emph{Journal of Machine Learning Research}, 18(13):1--43.

\bibitem[{Khodak et~al.(2018)Khodak, Saunshi, Liang, Ma, Stewart, and
  Arora}]{Khodak2018}
M.~Khodak, N.~Saunshi, Y.~Liang, T.~Ma, B.~Stewart, and S.~Arora. 2018.
\newblock A la carte embedding: Cheap but effective induction of semantic
  feature vectors.
\newblock \emph{To Appear in the Proceedings of the Association for Computation
  Linguistics (ACL)}.

\bibitem[{Mu et~al.(2017{\natexlab{a}})Mu, Bhat, and Viswanath}]{Mu2017b}
J.~Mu, S.~Bhat, and P.~Viswanath. 2017{\natexlab{a}}.
\newblock Representing sentences as low-rank subspaces.
\newblock In \emph{Proceedings of the 55th Annual Meeting of the Association
  for Computational Linguistics, {ACL} 2017, Vancouver, Canada, July 30 -
  August 4, Volume 2: Short Papers}, pages 629--634.

\bibitem[{Mu et~al.(2017{\natexlab{b}})Mu, Bhat, and Viswanath}]{Mu2017}
J.~Mu, S.~Bhat, and P.~Viswanath. 2017{\natexlab{b}}.
\newblock Representing sentences as low-rank subspaces.
\newblock In \emph{Proceedings of the 55th Annual Meeting of the Association
  for Computational Linguistics (Volume 2: Short Papers)}, pages 629--634.
  Association for Computational Linguistics.

\bibitem[{Mu and Viswanath(2018)}]{Mu2018}
J.~Mu and P.~Viswanath. 2018.
\newblock All-but-the-top: Simple and effective postprocessing for word
  representations.
\newblock In \emph{International Conference on Learning Representations}.

\bibitem[{Pennington et~al.(2014)Pennington, Socher, and
  Manning}]{Pennington2014}
J.~Pennington, R.~Socher, and C.~D. Manning. 2014.
\newblock Glove: Global vectors for word representation.
\newblock In \emph{Empirical Methods in Natural Language Processing (EMNLP)},
  pages 1532--1543.

\bibitem[{Stone et~al.(2011)Stone, Dennis, and Kwantes}]{Stone2011}
B.~Stone, S.~Dennis, and P.~J. Kwantes. 2011.
\newblock Comparing methods for single paragraph similarity analysis.
\newblock \emph{Topics in Cognitive Science}, 3(1):92--122.

\bibitem[{Wang et~al.(2018)Wang, Afzal, S.~Liu, Rastegar-Mojarad, Wang, Shen,
  and S.~Fu}]{Wang2018}
Y.~Wang, N.~Afzal, M~S.~Liu, Rastegar-Mojarad, L.~Wang, F.~Shen, and H.~Liu
  S.~Fu. 2018.
\newblock Overview of the biocreative/ohnlp challenge 2018 task 2: Clinical
  semantic textual similarity.
\newblock In \emph{Proceedings of the BioCreative/OHNLP Challenge. 2018}.

\bibitem[{Wieting et~al.(2016)Wieting, Bansal, Gimpel, and
  Livescu}]{Wieting2016}
J.~Wieting, M.~Bansal, K.~Gimpel, and K.~Livescu. 2016.
\newblock Towards universal paraphrastic sentence embeddings.
\newblock In \emph{International Conference on Learning Representations}.

\bibitem[{Wieting et~al.(2015)Wieting, Bansal, Gimpel, Livescu, and
  Roth}]{Wieting2015}
J.~Wieting, M.~Bansal, K.~Gimpel, K.~Livescu, and D.~Roth. 2015.
\newblock From paraphrase database to compositional paraphrase model and back.
\newblock \emph{Transactions of the Association for Computational Linguistics},
  3:345--358.

\bibitem[{Yin and Sch{\"u}tze(2015)}]{Yin2015}
W.~Yin and H.~Sch{\"u}tze. 2015.
\newblock Convolutional neural network for paraphrase identification.
\newblock In \emph{Proceedings of the 2015 Conference of the North American
  Chapter of the Association for Computational Linguistics: Human Language
  Technologies}, pages 901--911.

\end{thebibliography}


\end{document}